\documentclass[conference,a4paper]{IEEEtran}

\usepackage{amsmath}
\usepackage{booktabs}
\usepackage{graphicx}
\usepackage{cite}
\usepackage{url}
\usepackage{array}
\usepackage{amssymb}
\usepackage[table]{xcolor}

\newcommand{\anspass}[1]{\cellcolor{green!20}#1\,\(\checkmark\)}
\newcommand{\ansfail}[1]{\cellcolor{red!10}#1}
\newcommand{\finalok}[1]{#1}

\title{Context-Masked Truncated Reasoning Audits for Answer-Key Dependence in LLM Tutors}

\author{
\IEEEauthorblockN{Bonan Shen}
\IEEEauthorblockA{\textit{Independent Researcher}\\
United States\\
shenbonan2@gmail.com}
\and
\IEEEauthorblockN{Dingyan Shang}
\IEEEauthorblockA{\textit{Independent Researcher}\\
United States\\
dingyanshang@gmail.com}
\and
\IEEEauthorblockN{Youting Wang}
\IEEEauthorblockA{\textit{Northeastern University}\\
Boston, MA, USA\\
ginkoin613@gmail.com}
\and
\IEEEauthorblockN{Tao Ning}
\IEEEauthorblockA{\textit{Syracuse University}\\
Syracuse, NY, USA\\
ntgd1102@gmail.com}
\and
\IEEEauthorblockN{Bowen Liu}
\IEEEauthorblockA{\textit{Independent Researcher}\\
California, USA\\
bliu0962@usc.edu}
}

\begin{document}
\maketitle

\begin{abstract}
Large language model (LLM) tutors may have access to teacher notes, answer keys, rubrics, or retrieved solutions while producing student-facing explanations. We study whether truncated reasoning probes can distinguish direct access to such private context from answer information carried by the written explanation. Using Truncated Reasoning AUC Evaluation (TRACE), we evaluate 1000 GSM8K problems under question-only, correct answer-key, and wrong answer-key contexts. When forced-answer probes retain the private key, answer-key TRACE AUC rises from 0.375 to 0.900, and the gold answer is recoverable with no explanation at all in 998 of 1000 cases. We then introduce a context-masked replay: answer-key-generated prefixes are probed under the corresponding question-only prompt. Masking reduces 10\% prefix accuracy from 0.997 to 0.126 and median AUC from 0.900 to 0.375, nearly matching question-only values of 0.113 and 0.375. On 746 pairs where both explanations end correctly, the masked mean AUC difference is $-0.0086$ with a 95\% bootstrap interval spanning zero. Wrong keys still account for 272 of 387 incorrect final responses, showing that private artifacts can influence outputs even when early-prefix evidence disappears after masking. These results establish context masking as necessary for attributing early answer availability to an explanation rather than its hidden input.
\end{abstract}


\section{Introduction}

LLM tutors are increasingly asked to provide step-by-step help rather than only final answers. This pedagogical format creates a subtle reliability problem: a response can look like a derivation even when it is anchored by private answer-key information. Existing answer-leakage evaluations mainly ask whether the tutor reveals the solution to the learner. We ask a measurement-level counterpart: can a behavioral audit determine whether early answer availability comes from the explanation or from private context that remains visible to the probe?

Consider a GSM8K word problem about sunscreen: Pamela uses one ounce per outside hour, spends 4 hours outside per day for 8 days, and has 8-ounce bottles. A question-only explanation ends with the correct answer, 4, but its truncated prefixes fail until the full explanation. With a private answer key retained during probing, every answer-key prefix passes. Yet replaying those same answer-key-generated prefixes without the private key again produces no pass until the full explanation. The apparent early signal therefore came from the probe context, not the prefix. This example motivates our central question: when a truncated probe answers early, is the answer supported by the explanation or merely still visible in private input?

\begin{table}[t]
\centering
\caption{Motivating examples. ``AK, visible'' retains the answer key during each probe; ``AK, masked'' replays the same answer-key-generated prefix under the public question-only context. Green checkmarked cells match the gold answer.}
\label{tab:cases}
\scriptsize
\setlength{\tabcolsep}{2.0pt}
\renewcommand{\arraystretch}{1.12}

\newcommand{\problemrow}[2]{%
  \midrule
  \multicolumn{8}{@{}p{\linewidth}@{}}{\textbf{#1}. #2}\\[-1pt]
}

\begin{tabular}{@{}p{0.86in}>{\centering\arraybackslash}p{0.24in}>{\centering\arraybackslash}p{0.34in}*{5}{>{\centering\arraybackslash}p{0.33in}}@{}}
\toprule
Context & Final & AUC & \multicolumn{5}{c}{Forced answer at CoT prefix} \\
\cmidrule(l){4-8}
 & & & 10\% & 25\% & 50\% & 75\% & 100\% \\
\problemrow{Sunscreen, gold = 4}{Abridged: Pamela uses one ounce of sunscreen per outside hour. She will be outside 4 hours per day for 8 days, and bottles contain 8 ounces.}
Question-only & \finalok{4} & 0.125 & \ansfail{--} & \ansfail{8} & \ansfail{2} & \ansfail{5} & \anspass{4} \\
AK, visible & \finalok{4} & 0.900 & \anspass{4} & \anspass{4} & \anspass{4} & \anspass{4} & \anspass{4} \\
AK, masked & \finalok{4} & 0.125 & \ansfail{--} & \ansfail{8} & \ansfail{--} & \ansfail{5} & \anspass{4} \\
\problemrow{Flower soil, gold = 3}{Abridged: Artemis has 30 pounds of soil. Sunflowers need 3 pounds each, carnations need 1.5 pounds each, and roses need 1 pound each. They plant 4 sunflowers and 10 carnations.}
Question-only & \finalok{3} & 0.125 & \ansfail{10} & \ansfail{12} & \ansfail{11} & \ansfail{9} & \anspass{3} \\
AK, visible & \finalok{3} & 0.900 & \anspass{3} & \anspass{3} & \anspass{3} & \anspass{3} & \anspass{3} \\
AK, masked & \finalok{3} & 0.375 & \ansfail{6} & \ansfail{10} & \ansfail{14} & \anspass{3} & \anspass{3} \\
\problemrow{Checks/year, gold = 52}{Abridged: Buford writes monthly electric and gas checks, two church checks per month, and quarterly pest/lawn service checks.}
Question-only & \finalok{52} & 0.125 & \ansfail{18} & \ansfail{18} & \ansfail{54} & \ansfail{48} & \anspass{52} \\
AK, visible & \finalok{52} & 0.900 & \anspass{52} & \anspass{52} & \anspass{52} & \anspass{52} & \anspass{52} \\
AK, masked & \finalok{52} & 0.625 & \ansfail{13} & \ansfail{18} & \anspass{52} & \anspass{52} & \anspass{52} \\
\bottomrule
\end{tabular}

\end{table}

We adapt TRACE-style truncated reasoning evaluation \cite{wang2026trace} to educational math tutoring. TRACE measures when a reasoning prefix becomes sufficient to pass a verifier. In our setting, however, the candidate shortcut is part of the generation prompt itself. A probe that retains that prompt measures answer availability to the whole system, not sufficiency of the written prefix. We therefore distinguish a \emph{key-visible} deployment audit from a \emph{key-masked} explanation audit. Both use exact match to the GSM8K gold answer; neither judges whether the prose sounds pedagogical.

This paper makes three contributions. First, we identify a context-retention confound in truncated audits of tutors with private answer artifacts. Second, we introduce a paired context-masked replay that holds the generated prefix fixed while removing the private input during verification. Third, on 1000 GSM8K problems, we show that a large key-visible TRACE separation largely disappears after masking, while wrong-key interventions still demonstrate substantial final-output anchoring.

\section{Method}

Let $x_i$ be a math word problem with gold answer $y_i$. For each problem, we generate tutor explanations under three conditions:
\begin{itemize}
    \item \textbf{Question-only}: the model sees only the student-facing problem.
    \item \textbf{Answer-key}: the model sees the problem plus a teacher-only note containing the correct final answer.
    \item \textbf{Wrong-key}: the model sees the problem plus a teacher-only note containing a deterministic wrong answer, $y_i+1$.
\end{itemize}
All prompts ask the model to explain step by step and conclude with a boxed integer answer. The answer-key and wrong-key conditions instruct the model to use the key only as private context and write the explanation as if deriving the answer from the problem.

For a generated explanation containing $T$ tokens, we evaluate prefixes at fractions
\[
\mathcal{F}=\{0.10,0.25,0.50,0.75,1.00\}.
\]
At each prefix $fT$, we append a forced-answer tag asking the model to stop reasoning and give only the final answer. Let $p(f)$ be the verifier pass rate: the fraction of forced answers that equal the gold answer after integer normalization. In the main 1000-problem run, we use one deterministic forced answer per prefix, so $p(f)\in\{0,1\}$. These forced-prefix pass rates are distinct from original full-response accuracy: even at $f=1.00$, the probe asks for a fresh forced answer appended after the generated explanation, so the TRACE curve is not constrained to be monotone or identical to the original final answer. An early pass therefore indicates that the answer is recoverable from that prefix, not that all longer prefixes will preserve the same answer.

For ordered checkpoints $f_1,\ldots,f_m$, TRACE AUC is the trapezoidal area
\begin{equation}
\operatorname{AUC}=\sum_{j=1}^{m-1}(f_{j+1}-f_j)\frac{p(f_j)+p(f_{j+1})}{2}.
\end{equation}
Because the first checkpoint is 0.10, the reported AUC ranges from 0 to 0.90; an explanation that passes at every checkpoint has AUC 0.90. Pass@10 is $p(0.10)$, the earliest-prefix pass indicator. First-pass is the first prefix fraction where $p(f)\geq 0.5$; traces with no observed pass are right-censored at 1.0. Higher TRACE AUC and Pass@10 indicate earlier gold-answer availability, while lower first-pass indicates earlier first recoverability.

\section{Experimental Setup}

\textbf{Dataset.} We use 1000 examples from the GSM8K test split \cite{cobbe2021gsm8k}. GSM8K is suitable because it contains grade-school word problems with short numeric answers and natural solution structure. We use exact integer answer matching after comma and decimal normalization.

\textbf{Model and inference.} All experiments use Qwen2.5-3B-Instruct \cite{qwen2025technical} in 4-bit MLX format on an Apple M1 Max with 64 GB memory. Full explanations are greedily generated with a maximum of 768 tokens. Forced prefix answers use deterministic decoding with a maximum of 8 tokens. The run covers 1000 problems and 3 prompt conditions, yielding 3000 rows. A CUDA/Hugging Face runner was also prepared for replication, but the reported results use the local MLX run.

The system prompt asks the model to act as a careful grade-school math tutor, explain the solution step by step from the problem statement, and conclude with an integer in \verb|\boxed{X}| form. Keyed conditions append: ``Teacher-only answer key: the correct final answer is $a$. Use this only as private context. Write a tutor-style explanation as if deriving the answer from the problem,'' where $a=y$ or the counterfactual $y+1$. The forced-answer probe appends \verb|</think>| and asks the model to stop reasoning and return only a boxed final answer.

\textbf{Evaluation.} The unit of comparison is paired by problem. We compare the answer-key condition against question-only for the same problem, controlling for difficulty and answer value. We also report a subset in which both responses end correctly and a length-matched subset whose token counts differ by at most 10\%. Confidence intervals for mean paired differences use 20,000 problem-level bootstrap resamples with a fixed seed; wins, ties, and losses are computed before aggregation.

\textbf{Masked-context ablation.} In the primary audit, each forced completion retains the generation context, including any private key. This measures when the deployed tutor can produce the answer, but it does not isolate information carried by the written prefix. We therefore replay every saved answer-key prefix under the corresponding question-only prompt, removing the private key while holding the model, prefix text, checkpoints, and deterministic decoding fixed. We also probe at a 0\% prefix with and without the key. The masked answer-key curve can then be compared with the original question-only curve under the same public context; the unmasked 0\% probe quantifies direct answer availability from the private input alone.

This design yields two paired estimands. Let $A_i^{\mathrm{vis}}$ and $A_i^{\mathrm{mask}}$ be the AUCs of answer-key trace $i$ with the key visible and masked, and let $A_i^{Q}$ be its question-only counterpart. We define
\begin{align}
\Delta_{\mathrm{context},i} &= A_i^{\mathrm{vis}}-A_i^{\mathrm{mask}},\\
\Delta_{\mathrm{prefix},i} &= A_i^{\mathrm{mask}}-A_i^{Q}.
\end{align}
The first isolates answer availability supplied by retained private context. The second asks whether an answer-key-generated explanation exposes gold-answer evidence earlier than a question-only explanation once both are evaluated with the same public information. A large $\Delta_{\mathrm{context}}$ without a large $\Delta_{\mathrm{prefix}}$ indicates context access rather than prefix-level evidence.

\section{Results}

\subsection{Key-Visible Audit}

Table~\ref{tab:cases} illustrates the context-retention effect. All three answer-key explanations end correctly. With the key visible to the forced completion, every measured prefix passes. Once the key is removed, the sunscreen trace behaves exactly like question-only, the flower-soil trace first passes at 75\%, and the checks-per-year trace first passes at 50\%. Thus identical answer-key-generated text can yield qualitatively different curves depending on whether the probe can still read the private input.

\begin{table}[t]
\centering
\caption{Primary key-visible results. Orig. Acc. is unmodified tutor-response accuracy. AUC$_{50}$ and First$_{50}$ are medians; Pass@10 is a mean. Higher AUC and Pass@10 indicate earlier answer availability to the complete tutor context.}
\label{tab:main}
\begin{tabular}{lrrrrr}
\toprule
Condition & $n$ & Orig. Acc. & AUC$_{50}$ & Pass@10 & First$_{50}$ \\
\midrule
Question-only & 1000 & 0.782 & 0.375 & 0.113 & 0.750 \\
Answer-key & 1000 & 0.891 & 0.900 & 0.997 & 0.100 \\
Wrong-key & 1000 & 0.613 & 0.125 & 0.051 & 1.000 \\
\bottomrule
\end{tabular}

\end{table}

Table~\ref{tab:main} reproduces the large key-visible separation. Median AUC rises from 0.375 for question-only to 0.900 for answer-key, while Pass@10 rises from 0.113 to 0.997. The result establishes that the answer is available to a tutor that retains the key, but it cannot locate that information in the explanation because the forced completion also sees the key.

\begin{figure}[t]
\centering
\includegraphics[width=\columnwidth]{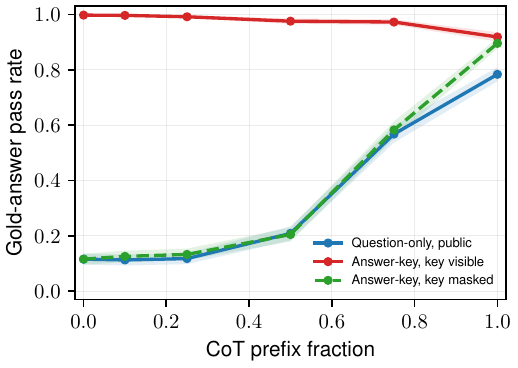}
\caption{Mean pass curves over 1000 problems. Retaining the key yields near-ceiling performance even at 0\%. Replaying answer-key prefixes under public context makes the curve closely track question-only. Bands are approximate 95\% confidence intervals for mean pass rates.}
\label{fig:curves}
\end{figure}

\subsection{Context-Masked Ablation}

Figure~\ref{fig:curves} and Table~\ref{tab:masked} show the decisive ablation. An answer-key probe with no explanation passes in 0.998 of problems when the key remains visible, compared with 0.116 under public context. After masking, answer-key Pass@10 falls from 0.997 to 0.126, close to the question-only rate of 0.113. Median answer-key AUC falls from 0.900 to 0.375, exactly matching the question-only median. The masked curve separates only near the end: at 100\%, answer-key prefixes pass at 0.896 versus 0.784 for question-only, consistent with their higher original final accuracy.

\begin{table}[t]
\centering
\caption{Mean verifier pass rates with and without private context. AUC$_{50}$ is the median over the 10--100\% grid; the 0\% probe is not included in AUC.}
\label{tab:masked}
\scriptsize
\setlength{\tabcolsep}{2.2pt}
\begin{tabular}{@{}llrrrrrrr@{}}
\toprule
Trace & Probe & 0\% & 10\% & 25\% & 50\% & 75\% & 100\% & AUC$_{50}$ \\
\midrule
Question-only & Public & 0.116 & 0.113 & 0.118 & 0.209 & 0.568 & 0.784 & 0.375 \\
Answer-key & Key visible & 0.998 & 0.997 & 0.992 & 0.976 & 0.973 & 0.919 & 0.900 \\
Answer-key & Key masked & 0.116 & 0.126 & 0.133 & 0.205 & 0.583 & 0.896 & 0.375 \\
\bottomrule
\end{tabular}

\end{table}

The original key-visible mean paired AUC difference is 0.551 (95\% bootstrap CI $[0.536,0.566]$). After masking, it is 0.0207 ($[0.0089,0.0330]$), with median 0: answer-key traces win 283 comparisons, tie 501, and lose 216. The small positive mean is concentrated near complete explanations rather than early prefixes.

Directly comparing each answer-key trace before and after masking gives mean $\Delta_{\mathrm{context}}=0.5303$ ($[0.5158,0.5444]$) and median 0.525; masking lowers AUC in 938 problems, leaves 60 tied, and raises it in only 2. At 0\%, removing the key lowers mean pass rate by 0.882 ($[0.861,0.902]$). These within-trace contrasts locate most of the original separation in retained context rather than generated text.

\subsection{Paired Robustness}

\begin{table}[t]
\centering
\caption{Paired answer-key (AK) versus question-only (Q) AUC comparisons. AK$>$Q, Tie, and AK$<$Q are percentages. ``Both correct'' fixes final-answer correctness; ``Length matched'' retains pairs whose explanation lengths differ by at most 10\%. $\Delta$AUC$_{50}$ is the median paired difference.}
\label{tab:paired}
\scriptsize
\begin{tabular}{@{}llrrrrr@{}}
\toprule
Probe & Subset & $n$ & AK$>$Q & Tie & AK$<$Q & $\Delta$AUC$_{50}$ \\
\midrule
Key visible & All & 1000 & 94.8 & 4.8 & 0.4 & 0.525 \\
Key visible & Both correct & 746 & 93.4 & 6.4 & 0.1 & 0.525 \\
Key masked & All & 1000 & 28.3 & 50.1 & 21.6 & 0.000 \\
Key masked & Both correct & 746 & 18.2 & 59.2 & 22.5 & 0.000 \\
Key masked & Length matched & 365 & 24.4 & 57.3 & 18.4 & 0.000 \\
\bottomrule
\end{tabular}

\end{table}

Table~\ref{tab:paired} confirms that the key-visible dominance does not survive context equalization. On the 746 pairs where both explanations end correctly, the masked mean difference is $-0.0086$ with 95\% CI $[-0.0201,0.0029]$; 442 pairs tie, 136 favor answer-key, and 168 favor question-only. Explanation length is also insufficient to explain the result. Median lengths are 329 tokens for question-only and 356 for answer-key. Among 365 pairs within 10\% length, the masked median difference remains 0 and the mean is 0.0169 ($[0.0001,0.0340]$).

\subsection{Where Does the Masked Difference Remain?}

Masking does not make the answer-key and question-only traces textually identical; it makes their early behavioral profiles similar. Their complete five-checkpoint pass vectors are identical on 493 problems, and their AUCs tie on 501. Table~\ref{tab:firstpass} shows nearly equal counts of first passes at 10\%, 25\%, 50\%, and 75\%. The main aggregate difference occurs at the boundary: masked answer-key traces have 277 first passes at 100\% and 84 with no pass, versus 211 and 167 for question-only. This is consistent with correct keys improving completed explanations rather than exposing the answer at their beginning.

\begin{table}[t]
\centering
\caption{Counts by first successful measured prefix. ``None'' means that all five forced completions fail.}
\label{tab:firstpass}
\scriptsize
\begin{tabular}{@{}lrrrrrr@{}}
\toprule
Trace & 10\% & 25\% & 50\% & 75\% & 100\% & None \\
\midrule
Question-only & 113 & 51 & 113 & 345 & 211 & 167 \\
Answer-key, masked & 126 & 48 & 115 & 350 & 277 & 84 \\
\bottomrule
\end{tabular}

\end{table}

The both-correct subset sharpens this interpretation. Relative to question-only, masked answer-key pass rates differ by only $+1.3$, $+0.8$, $-1.7$, $-2.8$, and $+0.1$ percentage points at 10\%, 25\%, 50\%, 75\%, and 100\%, respectively. By contrast, retaining the key creates 871 cases that pass at 10\% and fail after masking, with no cases in the reverse direction. Thus the dominant early-prefix asymmetry is specifically attributable to context visibility.

\begin{figure}[t]
\centering
\includegraphics[width=\columnwidth]{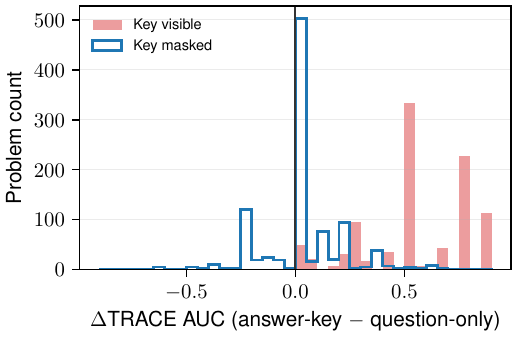}
\caption{Paired AUC differences before and after masking. The key-visible distribution is strongly positive; the masked distribution is centered at zero.}
\label{fig:delta}
\end{figure}

\subsection{Wrong-Key Counterfactual}

Masking removes evidence for unusually early gold-answer availability in answer-key-generated prefixes, but it does not imply that keys are behaviorally irrelevant. Replacing the key by $y+1$ lowers final accuracy from 0.782 in question-only to 0.613. Of 387 incorrect wrong-key responses, 272 (70.3\%) end with the injected value itself; another 113 produce other wrong answers and 2 contain no parsed answer. This counterfactual shows causal anchoring of final outputs, while the masked audit shows that such influence should not be attributed to an early prefix merely because a forced completion can still inspect the key.

TRACE curves also need not be monotone. Each point is a fresh deterministic completion conditioned on a different prefix, and later text can introduce an arithmetic error that overrides an earlier signal. In the key-visible condition, 884 traces pass at all five measured fractions, while 80 pass at 10\% but fail at 100\%. We therefore interpret individual points as conditional answer recoverability, not as cumulative proof that all later prefixes preserve an answer.

\section{Discussion}

The experiment separates three claims that a single truncated curve can conflate. First, a key-visible probe establishes \emph{system-level availability}: the tutor can answer because its full context contains the key. Second, the wrong-key intervention establishes \emph{behavioral dependence}: changing the private artifact changes many final outputs. Third, only a key-masked probe addresses \emph{prefix evidence}: whether the generated explanation itself makes the answer recoverable earlier. Our results strongly support the first two claims but provide little support for the third once correctness is controlled.

\subsection{Audit Workflow}

The proposed audit can be used as a lightweight diagnostic layer for educational LLM systems. First generate paired explanations with and without the private artifact. For a deployment-level question---whether the complete tutor can produce the answer---retain its original context during probing. For an explanation-level attribution, replay every prefix under a shared public context, include a 0\% baseline, and compare paired curves. A wrong-key or perturbed-artifact condition separately tests whether the private channel changes outputs.

This workflow is not a replacement for pedagogical evaluation. It does not score whether an explanation is clear, supportive, or appropriately scaffolded, and a masked pass does not reveal hidden model computation. It instead prevents a basic attribution error: crediting the written prefix for information that remains directly readable elsewhere in the prompt. The distinction applies beyond answer keys to retrieved solutions, grader feedback, rubrics, and tool outputs.

\subsection{Operational Interpretation}

The three tests support different deployment decisions. A high key-visible 0\% or early-prefix rate paired with a public-context baseline after masking indicates direct private-context availability. This may be acceptable for a teacher-facing assistant but is a warning sign when the system is expected to construct an independent derivation. A high masked early-prefix rate indicates that the generated text itself carries answer evidence early; such traces merit inspection for premature disclosure, post-hoc rationalization, or retrieval snippets copied into the response. Finally, a wrong-key intervention that changes final answers establishes behavioral dependence even when masked prefixes look ordinary. That pattern calls for retrieval validation or confidence-aware handling of teacher artifacts rather than a claim about early written reasoning.

Concurrent work on guarded explanation generation fixes the upstream decision and validates whether the generated text preserves it; DACRI demonstrates such a validation-and-fallback design in a supply-chain setting \cite{huang2026dacri}. This output-level safeguard addresses decision preservation rather than the source of evidence targeted by our context-masked audit.

These diagnostics should be reported separately instead of collapsed into one score. For model comparison, the public-context curve measures ordinary problem-solving progress, $\Delta_{\mathrm{context}}$ measures the contribution of retained private input, and $\Delta_{\mathrm{prefix}}$ measures additional evidence carried by answer-key-generated text. For system monitoring, thresholds can be calibrated on a trusted question-only set and stratified by final correctness and response length. Manual review can then focus on the smaller set of traces whose masked early-prefix behavior exceeds the baseline, rather than flagging every response produced in the presence of a key.

\textbf{Limitations.} Several validity boundaries remain. \emph{Construct validity}: context masking is an intervention, so an answer-key-generated prefix may be less natural under a public prompt. Holding the text fixed is precisely what isolates its informational contribution, but the result should not be interpreted as the model's unobserved internal state. The 0\% probe can solve some GSM8K questions directly; it is therefore a public-question baseline, not an empty-input baseline. \emph{External validity}: we use one 3B model, one numeric math dataset, a synthetic $y+1$ wrong key, and deterministic completions. Multi-turn tutoring, open-ended answers, and retrieved documents require richer perturbations and verifiers. \emph{Measurement validity}: exact-match parsing ignores pedagogical quality, and independently generated probe completions can be nonmonotone. Future work should test multiple model families, absolute-token checkpoints, stochastic probe replication, and human judgments of whether masked prefixes genuinely justify their answers.

\section{Related Work}

\textbf{Chain-of-thought reasoning and faithfulness.} Transformer language models \cite{vaswani2017attention} can be prompted to produce explicit reasoning traces, and chain-of-thought prompting improves performance on arithmetic and symbolic tasks \cite{wei2022cot,kojima2022zeroshot}. Decoding and decomposition methods such as self-consistency and program-aided reasoning further improve reasoning reliability \cite{wang2023selfconsistency,gao2023pal}. However, generated explanations are not guaranteed to be faithful to the model's actual decision process. Prior work formalizes faithfulness concerns \cite{jacovi2020faithfulness}, shows that CoT can rationalize biased or shortcut-driven answers \cite{turpin2023unfaithful}, and proposes intervention-based, information-flow, unlearning, counterfactual, and instance-level tests for faithfulness \cite{lanham2023faithfulness,paul2024making,li2025bettercot,tutek2025unlearning,balasubramanian2025closer,arcuschin2026wild,han2026rfeval,shen2025faithcot,shen2026cie}. Our work is aligned with this concern but focuses on a concrete educational setting where private answer artifacts may be present.

\textbf{Verification and truncated reasoning audits.} Mathematical reasoning benchmarks and verifiers make it practical to evaluate more than final answers \cite{cobbe2021gsm8k,hendrycks2021math}. Process-supervision work similarly emphasizes intermediate reasoning quality \cite{uesato2022process,lightman2024step}. TRACE detects implicit reward hacking by measuring how early a truncated trace can pass a verifier \cite{wang2026trace}, while computational-graph evaluations target process properties not visible from final outputs \cite{zhao2026graph}. We use the same truncation primitive but study a setting where the suspected shortcut is explicitly present in the input. This requires context masking: otherwise early verification can measure continued access to the shortcut rather than information exposed by the trace.

\textbf{AI tutor safety and leakage.} Recent tutoring evaluations study answer over-disclosure, scaffolding failures, pedagogical alignment, collaborative feedback, and robustness against adversarial students who try to extract answers \cite{macina2025mathtutorbench,dinucu2025teaching,yang2025collaborative,zhao2026leakage,hazra2026safetutors}. These works evaluate what the tutor reveals or how safely it teaches. We study a complementary attribution problem: whether a process probe can distinguish evidence exposed by the explanation from an answer artifact retained elsewhere in the tutor's context.

\textbf{Alignment and shortcut use.} Our setting is also related to broader concerns about proxy objectives and reward gaming \cite{skalse2022rewardgaming}. Empirical work on reward overoptimization shows how optimizing proxy feedback can degrade true task quality \cite{gao2023rewardoveropt}. We study a simpler educational analogue: the visible explanation may satisfy surface expectations while relying on an unintended answer source.

\section{Conclusion}

We introduced context-masked truncated reasoning audits for LLM tutors with private answer artifacts. On 1000 GSM8K problems, retaining the key produced a near-ceiling curve even with no explanation, whereas replaying the same answer-key-generated prefixes under public context reduced early performance to approximately the question-only baseline. Wrong keys nevertheless caused substantial final-answer anchoring. Truncated probes can therefore measure useful but different properties: key-visible probes test system-level answer availability, intervention tests establish artifact dependence, and context-masked probes test evidence carried by the explanation. Keeping these claims separate is necessary for defensible process-level auditing.

\IEEEtriggeratref{19}
\bibliographystyle{IEEEtran}
\bibliography{references}

@inproceedings{wang2026trace,
  title = {Is It Thinking or Cheating? Detecting Implicit Reward Hacking by Measuring Reasoning Effort},
  author = {Wang, Xinpeng and Joshi, Nitish and Plank, Barbara and Angell, Rico and He, He},
  booktitle = {International Conference on Learning Representations},
  year = {2026}
}

@inproceedings{wei2022cot,
  title = {Chain-of-Thought Prompting Elicits Reasoning in Large Language Models},
  author = {Wei, Jason and Wang, Xuezhi and Schuurmans, Dale and Bosma, Maarten and Ichter, Brian and Xia, Fei and Chi, Ed and Le, Quoc and Zhou, Denny},
  booktitle = {Advances in Neural Information Processing Systems},
  year = {2022}
}

@article{cobbe2021gsm8k,
  title = {Training Verifiers to Solve Math Word Problems},
  author = {Cobbe, Karl and Kosaraju, Vineet and Bavarian, Mohammad and Chen, Mark and Jun, Heewoo and Kaiser, Lukasz and Plappert, Matthias and Tworek, Jerry and Hilton, Jacob and Nakano, Reiichiro and Hesse, Christopher and Schulman, John},
  journal = {arXiv preprint arXiv:2110.14168},
  year = {2021}
}

@inproceedings{turpin2023unfaithful,
  title = {Language Models Don't Always Say What They Think: Unfaithful Explanations in Chain-of-Thought Prompting},
  author = {Turpin, Miles and Michael, Julian and Perez, Ethan and Bowman, Samuel R.},
  booktitle = {Advances in Neural Information Processing Systems},
  year = {2023}
}

@article{lanham2023faithfulness,
  title = {Measuring Faithfulness in Chain-of-Thought Reasoning},
  author = {Lanham, Tamera and Chen, Anna and Radhakrishnan, Ansh and Steiner, Benoit and Denison, Carson and Hernandez, Danny and Li, Dustin and Durmus, Esin and Hubinger, Evan and Kernion, Jackson and Luko{\v{s}}i{\={u}}t{\.e}, Kamil{\.e} and Nguyen, Karina and Cheng, Newton and Joseph, Nicholas and Schiefer, Nicholas and Rausch, Oliver and Larson, Robin and McCandlish, Sam and Kundu, Sandipan and Kadavath, Saurav and Yang, Shannon and Henighan, Thomas and Maxwell, Timothy and Telleen-Lawton, Timothy and Hume, Tristan and Hatfield-Dodds, Zac and Kaplan, Jared and Brauner, Jan and Bowman, Samuel R. and Perez, Ethan},
  journal = {arXiv preprint arXiv:2307.13702},
  year = {2023}
}

@inproceedings{zhao2026leakage,
  title = {Evaluating Answer Leakage Robustness of LLM Tutors against Adversarial Student Attacks},
  author = {Zhao, Jin and Kne{\v{z}}evi{\'c}, Marta and K{\"a}ser, Tanja},
  booktitle = {Proceedings of the 64th Annual Meeting of the Association for Computational Linguistics (Volume 1: Long Papers)},
  pages = {30588--30617},
  year = {2026}
}

@article{hazra2026safetutors,
  title = {SafeTutors: Benchmarking Pedagogical Safety in AI Tutoring Systems},
  author = {Hazra, Rima and Ghuku, Bikram and Marchenko, Ilona and Tokarieva, Yaroslava and Layek, Sayan and Banerjee, Somnath and Stoyanovich, Julia and Pechenizkiy, Mykola},
  journal = {arXiv preprint arXiv:2603.17373},
  year = {2026}
}

@inproceedings{vaswani2017attention,
  title = {Attention Is All You Need},
  author = {Vaswani, Ashish and Shazeer, Noam and Parmar, Niki and Uszkoreit, Jakob and Jones, Llion and Gomez, Aidan N. and Kaiser, Lukasz and Polosukhin, Illia},
  booktitle = {Advances in Neural Information Processing Systems},
  year = {2017}
}

@inproceedings{kojima2022zeroshot,
  title = {Large Language Models are Zero-Shot Reasoners},
  author = {Kojima, Takeshi and Gu, Shixiang Shane and Reid, Machel and Matsuo, Yutaka and Iwasawa, Yusuke},
  booktitle = {Advances in Neural Information Processing Systems},
  year = {2022}
}

@inproceedings{wang2023selfconsistency,
  title = {Self-Consistency Improves Chain of Thought Reasoning in Language Models},
  author = {Wang, Xuezhi and Wei, Jason and Schuurmans, Dale and Le, Quoc and Chi, Ed and Narang, Sharan and Chowdhery, Aakanksha and Zhou, Denny},
  booktitle = {International Conference on Learning Representations},
  year = {2023}
}

@inproceedings{gao2023pal,
  title = {PAL: Program-Aided Language Models},
  author = {Gao, Luyu and Madaan, Aman and Zhou, Shuyan and Alon, Uri and Liu, Pengfei and Yang, Yiming and Callan, Jamie and Neubig, Graham},
  booktitle = {International Conference on Machine Learning},
  year = {2023}
}

@article{uesato2022process,
  title = {Solving Math Word Problems with Process- and Outcome-Based Feedback},
  author = {Uesato, Jonathan and Kushman, Nate and Kumar, Ramana and Song, Francis and Siegel, Noah and Wang, Lisa and Creswell, Antonia and Irving, Geoffrey and Higgins, Irina},
  journal = {arXiv preprint arXiv:2211.14275},
  year = {2022}
}

@inproceedings{lightman2024step,
  title = {Let's Verify Step by Step},
  author = {Lightman, Hunter and Kosaraju, Vineet and Burda, Yura and Edwards, Harrison and Baker, Bowen and Lee, Teddy and Leike, Jan and Schulman, John and Sutskever, Ilya and Cobbe, Karl},
  booktitle = {International Conference on Learning Representations},
  year = {2024}
}

@inproceedings{hendrycks2021math,
  title = {Measuring Mathematical Problem Solving with the MATH Dataset},
  author = {Hendrycks, Dan and Burns, Collin and Kadavath, Saurav and Arora, Akul and Basart, Steven and Tang, Eric and Song, Dawn and Steinhardt, Jacob},
  booktitle = {NeurIPS Datasets and Benchmarks Track},
  year = {2021}
}

@inproceedings{jacovi2020faithfulness,
  title = {Towards Faithfully Interpretable NLP Systems: How Should We Define and Evaluate Faithfulness?},
  author = {Jacovi, Alon and Goldberg, Yoav},
  booktitle = {Annual Meeting of the Association for Computational Linguistics},
  year = {2020}
}

@inproceedings{skalse2022rewardgaming,
  title = {Defining and Characterizing Reward Gaming},
  author = {Skalse, Joar Max Viktor and Howe, Nikolaus H. R. and Krasheninnikov, Dmitrii and Krueger, David},
  booktitle = {Advances in Neural Information Processing Systems},
  year = {2022}
}

@inproceedings{gao2023rewardoveropt,
  title = {Scaling Laws for Reward Model Overoptimization},
  author = {Gao, Leo and Schulman, John and Hilton, Jacob},
  booktitle = {International Conference on Machine Learning},
  year = {2023}
}

@inproceedings{paul2024making,
  title = {Making Reasoning Matter: Measuring and Improving Faithfulness of Chain-of-Thought Reasoning},
  author = {Paul, Debjit and West, Robert and Bosselut, Antoine and Faltings, Boi},
  booktitle = {Findings of the Association for Computational Linguistics: EMNLP},
  pages = {15012--15032},
  year = {2024}
}

@inproceedings{li2025bettercot,
  title = {Towards Better Chain-of-Thought: A Reflection on Effectiveness and Faithfulness},
  author = {Li, Jiachun and Cao, Pengfei and Chen, Yubo and Xu, Jiexin and Li, Huaijun and Jiang, Xiaojian and Liu, Kang and Zhao, Jun},
  booktitle = {Findings of the Association for Computational Linguistics: ACL},
  pages = {10747--10765},
  year = {2025}
}

@inproceedings{tutek2025unlearning,
  title = {Measuring Chain of Thought Faithfulness by Unlearning Reasoning Steps},
  author = {Tutek, Martin and Hashemi Chaleshtori, Fateme and Marasovic, Ana and Belinkov, Yonatan},
  booktitle = {Proceedings of the 2025 Conference on Empirical Methods in Natural Language Processing},
  pages = {9935--9960},
  year = {2025}
}

@inproceedings{balasubramanian2025closer,
  title = {A Closer Look at Bias and Chain-of-Thought Faithfulness of Large (Vision) Language Models},
  author = {Balasubramanian, Sriram and Basu, Samyadeep and Feizi, Soheil},
  booktitle = {Findings of the Association for Computational Linguistics: EMNLP},
  pages = {13406--13439},
  year = {2025}
}

@inproceedings{arcuschin2026wild,
  title = {Chain-of-Thought Reasoning In The Wild Is Not Always Faithful},
  author = {Arcuschin, Iv{\'a}n and Janiak, Jett and Krzyzanowski, Robert and Rajamanoharan, Senthooran and Nanda, Neel and Conmy, Arthur},
  booktitle = {International Conference on Machine Learning},
  year = {2026}
}

@inproceedings{han2026rfeval,
  title = {RFEval: Benchmarking Reasoning Faithfulness under Counterfactual Reasoning Intervention in Large Reasoning Models},
  author = {Han, Yunseok and Lee, Yejoon and Do, Jaeyoung},
  booktitle = {International Conference on Learning Representations},
  year = {2026}
}

@inproceedings{zhao2026graph,
  title = {Verifying Chain-of-Thought Reasoning via Its Computational Graph},
  author = {Zhao, Zheng and Koishekenov, Yeskendir and Yang, Xianjun and Murray, Naila and Cancedda, Nicola},
  booktitle = {International Conference on Learning Representations},
  year = {2026}
}

@inproceedings{shen2025faithcot,
  title = {FaithCoT-Bench: Benchmarking Instance-Level Faithfulness of Chain-of-Thought Reasoning},
  author = {Shen, Xu and Wang, Song and Tan, Zhen and Yao, Laura and Zhao, Xinyu and Xu, Kaidi and Wang, Xin and Chen, Tianlong},
  booktitle = {International Conference on Learning Representations},
  year = {2026}
}

@article{shen2026cie,
  title = {Detecting Unfaithful Chain-of-Thought via Circuit-Guided Internal-External Discrepancy},
  author = {Shen, Xu and Tan, Zhen and Wang, Song and Hong, Pingjun and Miao, Rui and Wang, Xin and Chen, Tianlong},
  journal = {arXiv preprint arXiv:2605.25603},
  year = {2026}
}

@inproceedings{macina2025mathtutorbench,
  title = {MathTutorBench: A Benchmark for Measuring Open-ended Pedagogical Capabilities of LLM Tutors},
  author = {Macina, Jakub and Daheim, Nico and Hakimi, Ido and Kapur, Manu and Gurevych, Iryna and Sachan, Mrinmaya},
  booktitle = {Proceedings of the 2025 Conference on Empirical Methods in Natural Language Processing},
  pages = {204--221},
  year = {2025}
}

@inproceedings{dinucu2025teaching,
  title = {From Problem-Solving to Teaching Problem-Solving: Aligning LLMs with Pedagogy using Reinforcement Learning},
  author = {Dinucu-Jianu, David and Macina, Jakub and Daheim, Nico and Hakimi, Ido and Gurevych, Iryna and Sachan, Mrinmaya},
  booktitle = {Proceedings of the 2025 Conference on Empirical Methods in Natural Language Processing},
  pages = {272--292},
  year = {2025}
}

@inproceedings{yang2025collaborative,
  title = {LLM-based Collaborative Agents with Pedagogy-guided Interaction Modeling for Timely Instructive Feedback Generation in Task-oriented Group Discussions},
  author = {Yang, Qihao and Yang, Yu and An, Sixu and Hao, Tianyong and Xu, Guandong},
  booktitle = {Proceedings of the Thirty-Fourth International Joint Conference on Artificial Intelligence},
  pages = {9972--9980},
  year = {2025}
}

@article{qwen2025technical,
  title = {Qwen2.5 Technical Report},
  author = {{Qwen Team}},
  journal = {arXiv preprint arXiv:2412.15115},
  year = {2024}
}

@article{huang2026dacri,
  title = {{DACRI}: Decision-Aware Causal Intervention Ranking for Critical Supply Chains},
  author = {Huang, Shiqi and He, Jiani and Shang, Dingyan and Xu, Yihua and Li, Jize and Lyu, Yan and Nair, Lashimi Muraleedharan},
  journal = {arXiv preprint arXiv:2608.11154},
  year = {2026}
}

\end{document}